%% file: main.tex
\documentclass[10pt,twocolumn,letterpaper]{article}

\usepackage[pagenumbers]{cvpr} 

\input{preamble}

\definecolor{cvprblue}{rgb}{0.21,0.49,0.74}
\usepackage[pagebackref,breaklinks,colorlinks,citecolor=cvprblue]{hyperref}
\usepackage{amssymb}
\usepackage{pifont}
\newcommand{\cmark}{\ding{51}}%
\newcommand{\xmark}{\ding{55}}%
\usepackage{booktabs}
\usepackage{graphicx}
\def\paperID{*****} 
\def\confName{CVPR}
\def\confYear{2024}

\title{Unsupervised Multi-Person 3D Human Pose Estimation From 2D Poses Alone}

\author{Peter Hardy\\
University of Southampton\\
	Vision Learning and Control, ECS\\
{\tt\small p.t.d.hardy@soton.ac.uk}
\and
Hansung Kim Author\\
University of Southampton\\
	Vision Learning and Control, ECS\\
{\tt\small h.kim@soton.ac.uk}
}

\begin{document}
\maketitle
\input{sec/0_abstract}    
\input{sec/1_intro}
\input{sec/2_formatting}
\input{sec/3_finalcopy}
{
    \small
    \bibliographystyle{ieeenat_fullname}
    \bibliography{main}
}


\end{document}

%% file: preamble.tex
\usepackage[dvipsnames]{xcolor}


%% file: sec/0_abstract.tex
\begin{abstract}
Current unsupervised 2D-3D human pose estimation (HPE) methods do not work in multi-person scenarios due to perspective ambiguity in monocular images. Therefore, we present one of the first studies investigating the feasibility of unsupervised multi-person 2D-3D HPE from just 2D poses alone, focusing on reconstructing human interactions. To address the issue of perspective ambiguity, we expand upon prior work by predicting the cameras' elevation angle relative to the subjects' pelvis. This allows us to rotate the predicted poses to be level with the ground plane, while obtaining an estimate for the vertical offset in 3D between individuals. Our method involves independently lifting each subject's 2D pose to 3D, before combining them in a shared 3D coordinate system. The poses are then rotated and offset by the predicted elevation angle before being scaled. This by itself enables us to retrieve an accurate 3D reconstruction of their poses. We present our results on the CHI3D dataset, introducing its use for unsupervised 2D-3D pose estimation with three new quantitative metrics, and establishing a benchmark for future research.
\end{abstract}

%% file: sec/1_intro.tex
\section{Introduction}
\label{sec:intro}
\begin{figure}[t]
\centering
\includegraphics[width=0.95\columnwidth]{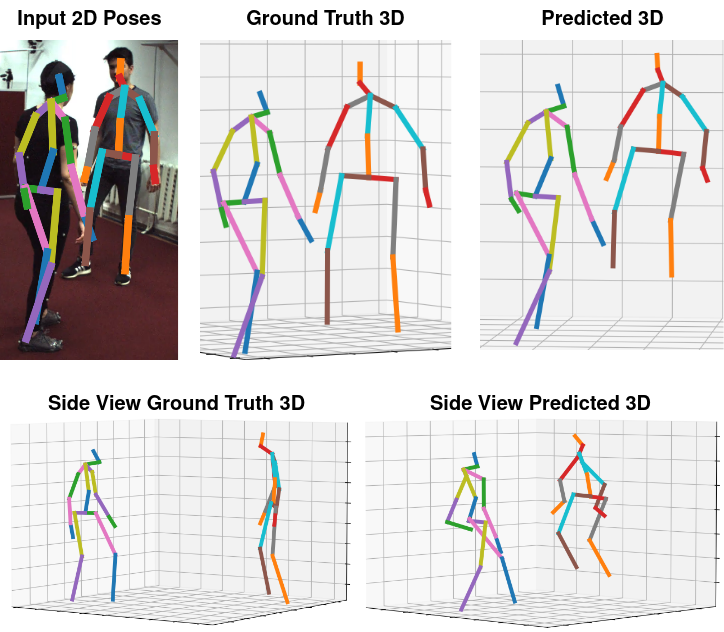}
\caption{Errors obtained when trying to use current unsupervised 2D-3D lifting approaches, to lift multiple people to 3D. In the above scenario, the root coordinate is the mid-point between each person's pelvis in 2D. We show a side view of the GT and Predicted 3D to highlight both pose prediction and 3D distance errors. Note how the person further back in the image appears to be floating and smaller in the predicted 3D when compared to the GT, this is due to the depth ambiguity in a perspective projection setting.}
\label{current_problems}
\end{figure}
\begin{figure*}[t]
\centering
\includegraphics[width=1.99\columnwidth]{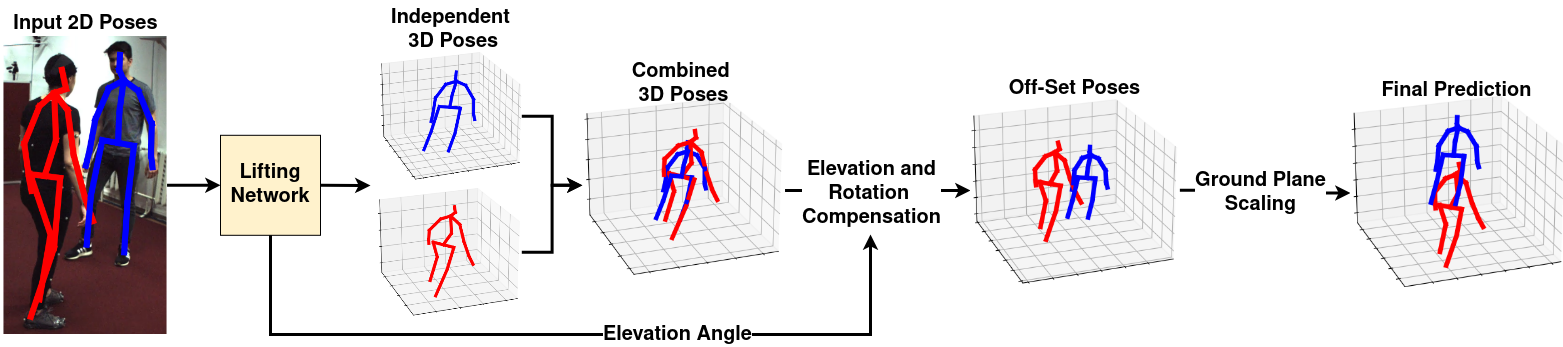}
\caption{Overview of our multi-person pose estimation approach. Given two or more detected 2D poses our lifting network \cite{hardy} predicts the 3D location for each joint for each pose independently. The 3D poses are then combined in their own global coordinate system. An elevation compensation approach accurately predicts the offset of each person's pelvis in a 3D setting. Lastly, each pose is scaled so that their feet are on the same ground plane which produces our final prediction.}
\label{end_to_end}
\end{figure*}
Monocular 3D human pose estimation (HPE) is known to be an ill-posed inverse problem, as multiple different 2D poses can correspond to the same 3D pose. Despite this, unsupervised algorithms have developed rapidly in tackling single-person 2D-3D HPE from a single image, where they attempt to lift a 2D skeleton to 3D via some form of reprojected 2D pose likelihood \cite{hardy, Wandt_2022_CVPR, amazon_paper_2, Yu_2021_ICCV, Drover_2018_ECCV_Amazon}. Due to fundamental perspective ambiguity however, the absolute metric of depth cannot be obtained from a single view alone \cite{amazon_paper_2, depth_impossible}. To deal with this, unsupervised approaches centre the detected 2D pose around a root joint (typically the pelvis), while also setting the 3D prediction of the root to be a fixed unit of $c$ from the camera. This means that instead of absolute depth, these models learn to predict the 3D depth offset from the root joint when the root is assumed to be $c$ units from the camera. Although this works well in scenarios where we want to lift a single person, if we adapt this approach to lift multiple people simultaneously we obtain errors both in terms of the pose, as well as in the 3D distance between the two people as seen in Fig. \ref{current_problems}. Therefore, the aim of this paper is to obtain an accurate reconstruction of 3D human poses interacting within a shared coordinate system, relying solely on their 2D poses obtained from a single image. Additionally, in our extensive survey of prior literature, we found no prior work tackling unsupervised multi-person 2D-3D HPE from 2D poses alone. Therefore, we take the first leap in exploring if it is feasible to reconstruct an accurate 3D estimate of two people interacting from their 2D poses alone. 

%% file: sec/2_formatting.tex
\section{Formatting your paper}
\label{sec:formatting}

\section{Methodology}
In this section, we present our unsupervised learning approach to independently lift 2D poses to 3D, combining them to a shared coordinate space and predicting the relative elevation angle of each person which is used for elevation and rotation compensation. An illustrative depiction of our approach is provided in Fig.\ref{end_to_end}. Our 2D poses consisted of $N$ keypoints, $(x_i, y_i)$, $i = 1...N$, where the root keypoint, located at the origin $(0, 0)$, was the midpoint between the left and right hip joint (pelvis). Similar to prior work we adopted the practice of fixing the distance of the pose from the camera at a constant $c$ units and normalising such that the average distance from the head to the root keypoint was $\frac{1}{c}$ units in 2D, with $c$ being set to 10 as is consistent with previous research \cite{amazon_paper_2, Wandt_2022_CVPR, Yu_2021_ICCV, hardy}. 

\subsection{Independent Lifting and Pose Combining}
Our lifting networks were trained to predict the 3D depth offset ($\hat{d}$) from the poses root keypoint for each 2D keypoint $(x, y)$. To compute the final 3D location of a specific keypoint, $\mathbf{x}_i$, we employed perspective projection, as defined by:
\begin{equation}
\begin{split}
\mathbf{x}_i &= (x_i\hat{z}_i, y_i\hat{z}_i, \hat{z}_i), \\
\mathbf{where } \quad \hat{z}_i &= \max(1, \hat{d}_i + c).
\end{split}
\end{equation}
Here, $d_i$ represents the depth-offset prediction made by our lifting network for keypoint $i$. Each 2D pose obtained from an image was lifted into 3D independently. This approach effectively mitigated the 3D pose estimation errors present when lifting both poses together, as demonstrated in Fig.\ref{current_problems}. Since both 3D poses shared the same root location, they were combined into a unified coordinate system, as depicted in the 'Combined 3D Poses' section in Fig.\ref{end_to_end}. For our lifting network, we adopted the LInKs algorithm, originally introduced by Hardy and Kim \cite{hardy}. We extended this algorithm to lift two additional keypoints, specifically the left and right hands. This inclusion was motivated by the significance of hand movements in contact-based interactions. It is worth noting that these additional keypoints are not typically present in most 3D pose datasets, such as Human3.6M \cite{h36m_pami} and MPI-INF-3DHP \cite{mono-3dhp2017}.

\subsection{3D Elevation and Rotation Compensation}
\begin{figure}[t]
\centering
\includegraphics[width=0.99\columnwidth]{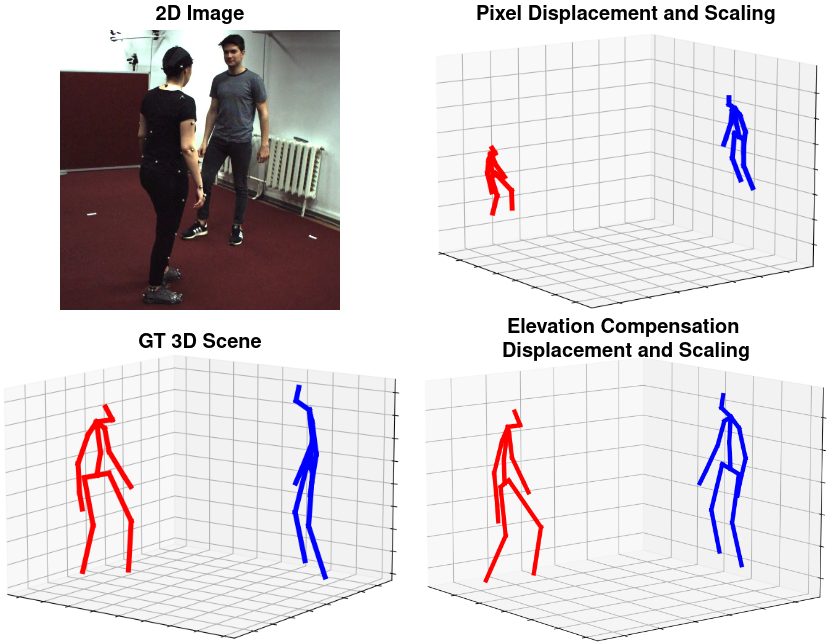}
\caption{Top right shows the errors obtained in both scaling and displacement when we assume that the original vertical 2D displacement of the poses (top left) accurately represents the height offset in the real world. The bottom right image shows our proposed elevation compensation approach to displacement and scaling, allowing for more accurate depth offset and scaling to be predicted.}
\label{displacement}
\end{figure}

As natural human behaviour places the subjects of interest in the centre of an image while holding the camera horizontally, and ``narrow-angle" lenses typically have little or no horizontal distortion \cite{Buquet_2021_CVPR, Wandt_2022_CVPR}, we can assume the horizontal displacement of the 2D poses can correspond to the horizontal displacement in 3D once scaled. However, if we naively used the elevation displacement in the 2D poses, and then applied scaling, we would obtain substantial errors, as depicted in Fig.\ref{displacement}. 
These errors predominantly stem from the variable elevation angle of the camera, which when angled up or down exaggerates the perceived height differences for people in a scene. Moreover, any predicted 3D pose will be tilted to or from the camera depending on the camera's elevation angle, further increasing the error. To tackle this, we expanded the work of Wandt \emph{et al.} \cite{Wandt_2022_CVPR}, who noticed that using a random elevation angle during 3D rotation and 2D reprojection can lead to unnatural 2D poses. Therefore, they sought to compensate by learning the elevation angle that would align the camera with the root of the pose before any further 3D transformations. However, it was not considered in their work how this predicted elevation angle could be used to calculate the elevation offset for two poses from the same scene. Therefore we include the same elevation angle prediction branch in our own lifting network while using the elevation angle prediction for the following additional steps. Let us have two predicted 3D poses, $\mathbf{y}_1$ and $\mathbf{y}_2$, along with their respective elevation angle predictions, $\theta_1$ and $\theta_2$. As the root of both poses is predicted to be $c$ units from the camera, the vertical distance the camera needs to move to align with the root of pose $\mathbf{y}_1$ and $\mathbf{y}_2$, is given by $c \cdot \tan(\theta_1)$ and $c \cdot \tan(\theta_2)$ respectively. Consequently, we estimated the vertical offset $\Delta h$ of $\mathbf{y}_1$ and $\mathbf{y}_2$ by considering the difference in vertical distance the camera has to move to align with the root of each pose:
\begin{equation}
\Delta h = c \cdot (\tan(\theta_1) - \tan(\theta_2))
\end{equation}
To solve the error of our 3D poses being tilted depending on the camera's elevation angle, we also introduced a rotational compensation for each pose around the $x$ axis. To do this we created the rotation matrix $\mathbf{R}_1$ and $\mathbf{R}_2$ from each poses respective $\theta$ where:
\begin{equation}
\mathbf{R} = \begin{bmatrix}
    1 & 0 & 0 \\
    0 & \cos(-\theta) & -\sin(\theta) \\
    0 & \sin(\theta) & \cos(\theta)
\end{bmatrix}
\end{equation}
In summary, we obtained the horizontal displacement of  $\mathbf{y}_1$ and $\mathbf{y}_2$ from the image, and the elevation displacement was obtained via trigonometry using the predicted elevation angle of each pose. First, as both poses are centred around the same origin, we rotated them by their respective rotation matrix $\mathbf{R}$. We then displaced both poses along the $x$ and $y$ axis, based on the horizontal distance in the image and the predicted vertical distance via elevation compensation. Lastly, we scaled each pose so that the lower of their two feet were aligned on the $y$ plane. This entire process led to the complete 3D reconstruction of the scene.

%% file: sec/3_finalcopy.tex
\section{Final copy}

\section{Evaluation}
The Close Human Interactions (CHI3D) dataset introduced by Fieraru \emph{et al.} \cite{Fieraru_2020_CVPR}, was one of the first 3D human interaction datasets, and new at the time of writing. It contains the 3D ground truth (GT) joints from mocap obtained from images taken by 4 cameras. Each sequence contains 2 people in various interactions such as grabbing, pushing, or holding hands. In our extensive literature review, we found that only two previous publications mentioned the CHI3D dataset in their writing. These publications include the original dataset release, and subsequent research work led by the same authors \cite{neurips_chi3d_eval_only}. Additionally, we found two other studies that submitted their results for evaluation on the CHI3D webpage \cite{li2022cliff, jiang2020mpshape}. All of these approaches relied on imagery or video, and estimated the GHUM \cite{GHUM} and SMPLX \cite{SMPLX} body models for evaluation. Furthermore, out of the 5 sequences that make up CHI3D the 3D data pertaining to sequences 1 and 5, as employed for evaluation in prior research \cite{neurips_chi3d_eval_only, Fieraru_2020_CVPR, li2022cliff, jiang2020mpshape}, has not been made publicly available at the time of writing. As we do not use imagery or videos in our study, but just 2D poses, we are unable to train or evaluate our approach using the same protocols. Therefore we first detail our training approach as well as the four evaluation metrics we use and their definitions, followed by our results on the CHI3D dataset.

\subsection{Training Approach and Error Metrics}
To train our lifting models and normalising flows we use the 2D pose data in sequences 2 and 3. Sequence 4 is then used for evaluation. As the relative size of the CHI3D dataset is much smaller than traditional HPE datasets, we do not use ``interesting" frames depending on the subjects' movement, but use all frames for training and evaluation. We pre-train our normalising flows for 100 epochs and train our lifting network for 40 epochs. We use the Adam Optimiser with an initial learning rate of $2\times10^{-4}$ which decayed exponentially by 0.95 every epoch. We used an identical architecture for our flows and lifting networks as detailed within Hardy and Kim \cite{hardy}. Additionally, as our predicted poses are within a normalised 3D coordinate system, we aligned them to the GT via Procrustes alignment prior to evaluation. Note that we treat the poses within our scene as one rigid structure during alignment, meaning that if pose 1 was scaled by $s$ and translated by $t$ the same would happen to pose 2. The evaluation metrics we used are:
\begin{itemize}
    \item \textbf{PA-MPJPE:} PA-MPJPE is the mean per joint position error in millimetres (mm), representing the Euclidean distance between the predicted and the GT 3D keypoints. Unlike prior approaches, we report this error collectively for all poses within a scene instead of for each pose individually. This inflates the error but provides a more accurate and comprehensive depiction of the joint errors within the reconstructed 3D scene.
    
    \item \textbf{Scale Error (SE):} SE represents the mean difference in mm between the L2 norm of the poses within our predicted and GT scenes. In other words, it assesses how much the total size of the poses in our predicted scene deviates from the poses in the GT scene. SE offers a detailed evaluation of scaling accuracy by focusing specifically on pose size instead of the overall scene size.

    \item \textbf{Translation Error (TE):} TE represents the L2 norm of the mean absolute translation error in mm between our predicted and GT 3D scenes. It provides insight into the accuracy of our 3D reconstruction with respect to translation.

    \item \textbf{Root Displacement Error (RDE):} RDE quantifies the mean error in mm between the pelvis displacement in the GT scene and the pelvis displacement in our predicted 3D scene. The RDE metric assesses whether our predicted 3D poses are displaced by the correct amount within our reconstruction. 
\end{itemize}

\subsection{Results and Limitations}
\begin{figure}[ht]
\centering
\includegraphics[width=0.95\columnwidth]{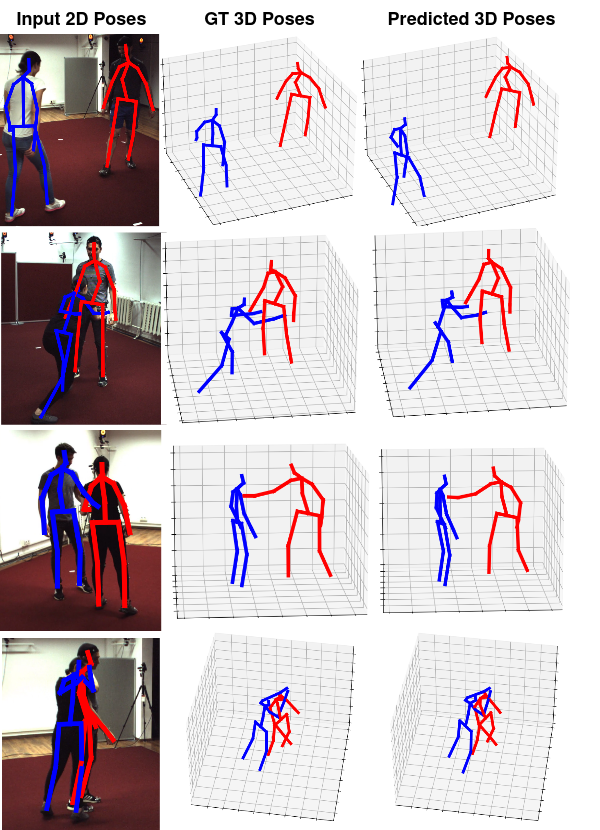}
\caption{Qualitative results on the CHI3D dataset}
\label{qualitative}
\end{figure}

We show the results of our model using 2D image displacement, our elevation compensation approach for displacement, and elevation and rotation compensation in Table \ref{tab:chi3d_results}. Qualitative results can be seen in Fig. \ref{qualitative}. Our results show that both of our changes improved results, with the rotation compensation alone reducing the PA-MPJPE error by 23.4\%. Furthermore, both of our changes reduced the error in scaling and displacement between the predicted and GT poses, showing how much the elevation angle of the camera can exaggerate the size and displacement of people within a scene. The main limitation of our approach is that it relies on an accurate 2D pose estimate to perform optimally, particularly an accurate pelvis keypoint. This is because the elevation angle prediction depends on this keypoint being accurate. Furthermore, we find multiple discrepancies in the CHI3D dataset between the 2D annotation, images and 3D poses. For example, when there is contact between subjects, the image and GT 3D poses showed a negligible vertical displacement in the pelvises along the $y$ axis. However, the 2D annotation often showed a large displacement. To mitigate this, when the vertical pelvis distance between both poses in the image was 50 pixels or less, we assumed $\theta_1$ = $\theta_2$. To mitigate this constraint, in future work we plan on combining our elevation compensation approach with a contact detector. This would allow us to use the contact point as a reference when displacing and scaling the poses removing this constraint.
\begin{table}[h]
\centering
\resizebox{\columnwidth}{!}{%
\begin{tabular}{@{}lccccccc@{}}
\toprule
Method &
  \begin{tabular}[c]{@{}c@{}}Elevation\\ Compensation\end{tabular} &
  \begin{tabular}[c]{@{}c@{}}Rotation\\ Compensation\end{tabular} &
  \begin{tabular}[c]{@{}c@{}}$\theta_1$ = $\theta_2$\\ when $\leq$ 50 pixels\end{tabular} &
  PA-MPJPE $\downarrow$ &
  SE $\downarrow$ &
  MTE $\downarrow$ &
  RDE $\downarrow$ \\ \midrule
Ours & \xmark     & \xmark     & \xmark     & 208.1 & $\pm$5.3\% & 87.6 & 148.4 \\
Ours & \xmark     & \xmark     & \cmark & 166.1 & $\pm$4.8\%  & 76.5 & 129.4 \\
Ours & \xmark     & \cmark & \cmark & 163.4 & $\pm$4.1\%  & 74.6 & 126.6 \\ \midrule
Ours & \cmark & \cmark & \cmark & 149.4 & $\pm$2.5\%  & 69.9 & 105.8 \\ \bottomrule
\end{tabular}%
}
\caption{Ablation results showing the effect of all of our changes for each of our error metrics on the CHI3D dataset.}
\label{tab:chi3d_results}
\end{table}

\section{Conclusion}
To conclude, we presented a novel and one of the first unsupervised approaches to reconstruct multi-person 3D human interactions from 2D poses alone. We identified the limitations in using 2D pose displacement in an image due to camera elevation angle variations and addressed this with an elevation and rotation compensation method. Our lifting and compensation approach distinguishes itself from other 3D reconstruction methods by its lightweight nature capable of real-time operation \cite{real_time}, thereby enhancing its applicability to real-world scenarios. Furthermore, our study is among the first to utilise the CHI3D dataset for 3D and introduced three new quantitative measures to demonstrate the accuracy of our approach and to serve as benchmarks for future research. We hope our work will inspire further research into the challenging task of unsupervised multi-person 3D reconstruction from 2D poses alone.

%% file: main.bbl
\begin{thebibliography}{16}
\providecommand{\natexlab}[1]{#1}
\providecommand{\url}[1]{\texttt{#1}}
\expandafter\ifx\csname urlstyle\endcsname\relax
  \providecommand{\doi}[1]{doi: #1}\else
  \providecommand{\doi}{doi: \begingroup \urlstyle{rm}\Url}\fi

\bibitem[Buquet et~al.(2021)Buquet, Zhang, Roulet, Thibault, and Lalonde]{Buquet_2021_CVPR}
Julie Buquet, Jinsong Zhang, Patrice Roulet, Simon Thibault, and Jean-Francois Lalonde.
\newblock Evaluating the impact of wide-angle lens distortion on learning-based depth estimation.
\newblock In \emph{Proceedings of the IEEE/CVF Conference on Computer Vision and Pattern Recognition (CVPR) Workshops}, pages 3693--3701, 2021.

\bibitem[Chen et~al.(2019)Chen, Tyagi, Agrawal, Drover, MV, Stojanov, and Rehg]{amazon_paper_2}
Ching-Hang Chen, Ambrish Tyagi, Amit Agrawal, Dylan Drover, Rohith MV, Stefan Stojanov, and James~M. Rehg.
\newblock Unsupervised 3d pose estimation with geometric self-supervision.
\newblock In \emph{2019 IEEE/CVF Conference on Computer Vision and Pattern Recognition (CVPR)}, pages 5707--5717, 2019.

\bibitem[Drover et~al.(2019)Drover, M.~V, Chen, Agrawal, Tyagi, and Huynh]{Drover_2018_ECCV_Amazon}
Dylan Drover, Rohith M.~V, Ching-Hang Chen, Amit Agrawal, Ambrish Tyagi, and Cong~Phuoc Huynh.
\newblock Can 3d pose be learned from 2d projections alone?
\newblock In \emph{Computer Vision -- ECCV 2018 Workshops}, pages 78--94, Cham, 2019. Springer International Publishing.

\bibitem[Fieraru et~al.(2020)Fieraru, Zanfir, Oneata, Popa, Olaru, and Sminchisescu]{Fieraru_2020_CVPR}
Mihai Fieraru, Mihai Zanfir, Elisabeta Oneata, Alin-Ionut Popa, Vlad Olaru, and Cristian Sminchisescu.
\newblock Three-dimensional reconstruction of human interactions.
\newblock In \emph{The IEEE/CVF Conference on Computer Vision and Pattern Recognition (CVPR)}, 2020.

\bibitem[Fieraru et~al.(2021)Fieraru, Zanfir, Szente, Bazavan, Olaru, and Sminchisescu]{neurips_chi3d_eval_only}
Mihai Fieraru, Mihai Zanfir, Teodor Szente, Eduard Bazavan, Vlad Olaru, and Cristian Sminchisescu.
\newblock Remips: Physically consistent 3d reconstruction of multiple interacting people under weak supervision.
\newblock In \emph{Advances in Neural Information Processing Systems}, pages 19385--19397. Curran Associates, Inc., 2021.

\bibitem[Hardy and Kim()]{hardy}
Peter Hardy and Hansung Kim.
\newblock {LInKs - Lifting Independent Keypoints - Exploring Partial Pose Lifting for Occlusion Handling and Improved Accuracy within 2D-3D Human Pose Estimation}.
\newblock In \emph{Submitted to WACV 2024 (Submission Version Available Online @ https://github.com/Aswarin/Papers}.

\bibitem[Ionescu et~al.(2014)Ionescu, Papava, Olaru, and Sminchisescu]{h36m_pami}
Catalin Ionescu, Dragos Papava, Vlad Olaru, and Cristian Sminchisescu.
\newblock Human3.6m: Large scale datasets and predictive methods for 3d human sensing in natural environments.
\newblock \emph{IEEE Transactions on Pattern Analysis and Machine Intelligence}, 36\penalty0 (7):\penalty0 1325--1339, 2014.

\bibitem[Jiang et~al.(2020)Jiang, Kolotouros, Pavlakos, Zhou, and Daniilidis]{jiang2020mpshape}
Wen Jiang, Nikos Kolotouros, Georgios Pavlakos, Xiaowei Zhou, and Kostas Daniilidis.
\newblock Coherent reconstruction of multiple humans from a single image.
\newblock In \emph{CVPR}, 2020.

\bibitem[Knap et~al.()Knap, Hardy, and Kim]{real_time}
Pawel Knap, Peter Hardy, and Hansung Kim.
\newblock {Real-time omnidirectional 3D multi-person human pose estimation with occlusion handling}.
\newblock In \emph{Submitted to CVMP 2023 (Submission Version Available Online @ https://github.com/Aswarin/Papers)}.

\bibitem[Li et~al.(2022)Li, Liu, Zhang, Xu, and Yan]{li2022cliff}
Zhihao Li, Jianzhuang Liu, Zhensong Zhang, Songcen Xu, and Youliang Yan.
\newblock Cliff: Carrying location information in full frames into human pose and shape estimation.
\newblock In \emph{ECCV}, 2022.

\bibitem[Mehta et~al.(2017)Mehta, Rhodin, Casas, Fua, Sotnychenko, Xu, and Theobalt]{mono-3dhp2017}
Dushyant Mehta, Helge Rhodin, Dan Casas, Pascal Fua, Oleksandr Sotnychenko, Weipeng Xu, and Christian Theobalt.
\newblock Monocular 3d human pose estimation in the wild using improved cnn supervision.
\newblock In \emph{3D Vision (3DV), 2017 Fifth International Conference on}. IEEE, 2017.

\bibitem[Nishimura et~al.(2020)Nishimura, Lindell, Metzler, and Wetzstein]{depth_impossible}
Mark Nishimura, David~B. Lindell, Christopher Metzler, and Gordon Wetzstein.
\newblock Disambiguating monocular depth estimation with a single transient.
\newblock In \emph{Computer Vision -- ECCV 2020}, pages 139--155, Cham, 2020. Springer International Publishing.

\bibitem[Pavlakos et~al.(2019)Pavlakos, Choutas, Ghorbani, Bolkart, Osman, Tzionas, and Black]{SMPLX}
Georgios Pavlakos, Vasileios Choutas, Nima Ghorbani, Timo Bolkart, Ahmed A.~A. Osman, Dimitrios Tzionas, and Michael~J. Black.
\newblock Expressive body capture: 3d hands, face, and body from a single image.
\newblock In \emph{Proceedings IEEE Conf. on Computer Vision and Pattern Recognition (CVPR)}, 2019.

\bibitem[Wandt et~al.(2022)Wandt, Little, and Rhodin]{Wandt_2022_CVPR}
Bastian Wandt, James~J. Little, and Helge Rhodin.
\newblock Elepose: Unsupervised 3d human pose estimation by predicting camera elevation and learning normalizing flows on 2d poses.
\newblock In \emph{Proceedings of the IEEE/CVF Conference on Computer Vision and Pattern Recognition (CVPR)}, pages 6635--6645, 2022.

\bibitem[Xu et~al.(2020)Xu, Bazavan, Zanfir, Freeman, Sukthankar, and Sminchisescu]{GHUM}
Hongyi Xu, Eduard~Gabriel Bazavan, Andrei Zanfir, William~T. Freeman, Rahul Sukthankar, and Cristian Sminchisescu.
\newblock Ghum \& ghuml: Generative 3d human shape and articulated pose models.
\newblock In \emph{2020 IEEE/CVF Conference on Computer Vision and Pattern Recognition (CVPR)}, pages 6183--6192, 2020.

\bibitem[Yu et~al.(2021)Yu, Ni, Xu, Wang, Zhao, and Zhang]{Yu_2021_ICCV}
Zhenbo Yu, Bingbing Ni, Jingwei Xu, Junjie Wang, Chenglong Zhao, and Wenjun Zhang.
\newblock Towards alleviating the modeling ambiguity of unsupervised monocular 3d human pose estimation.
\newblock In \emph{Proceedings of the IEEE/CVF International Conference on Computer Vision (ICCV)}, pages 8651--8660, 2021.

\end{thebibliography}
